\pdfoutput=1

\documentclass[11pt]{article}

\usepackage{acl}

\usepackage{times}
\usepackage[inline]{enumitem}

\usepackage{latexsym}

\usepackage[T1]{fontenc}

\usepackage[utf8]{inputenc}

\usepackage{microtype}

\usepackage{inconsolata}

\usepackage{verbatim}
\usepackage{flushend}
\usepackage{tabularx}
\usepackage{soul}
\usepackage{url}
\usepackage{booktabs}
\usepackage{multirow}
\usepackage{caption}
\usepackage{graphicx}
\usepackage{color}

\title{Beyond Text-to-SQL for IoT Defense: A Comprehensive Framework for Querying and Classifying IoT Threats}
\author{Ryan Pavlich$^1$, Nima Ebadi$^2$, Richard Tarbell$^1$, Billy Linares$^1$, Adrian Tan$^1$,\\\textbf{Rachael Humphreys$^1$, Jayanta Kumar Das$^1$, Rambod Ghandiparsi$^1$, Hannah Haley$^1$,}\\\textbf{Jerris George$^1$, $^3$Rocky Slavin, $^4$Kim-Kwang Raymond Choo,}\\\textbf{$^4$Glenn Dietrich,\and $^4$Anthony Rios} \\
  $^1$Data Analytics, $^2$Department of Electrical and Computer Engineering,\\$^3$Department of Computer Science, $^4$Department of Information Systems and Cyber Security\\
  The University of Texas at San Antonio\\
  \texttt{\{Ryan.Palvich, Anthony.Rios\}@utsa.edu} \\} 

\begin{document}

\maketitle

\begin{abstract}
Recognizing the promise of natural language interfaces to databases, prior studies have emphasized the development of text-to-SQL systems. While substantial progress has been made in this field, existing research has concentrated on generating SQL statements from text queries. The broader challenge, however, lies in inferring new information about the returned data. Our research makes two major contributions to address this gap. First, we introduce a novel Internet-of-Things (IoT) text-to-SQL dataset comprising 10,985 text-SQL pairs and 239,398 rows of network traffic activity.  The dataset contains additional query types limited in prior text-to-SQL datasets, notably temporal-related queries. Our dataset is sourced from a smart building's IoT ecosystem exploring sensor read and network traffic data. Second, our dataset allows two-stage processing, where the returned data (network traffic) from a generated SQL can be categorized as malicious or not. Our results show that joint training to query and infer information about the data can improve overall text-to-SQL performance, nearly matching substantially larger models. We also show that current large language models (e.g., GPT3.5) struggle to infer new information about returned data, thus our dataset provides a novel test bed for integrating complex domain-specific reasoning into LLMs.
\end{abstract}

\section{Introduction}

Relational databases contain vast quantities of structured knowledge, often having trillions of rows of data, spanning diverse domains from healthcare and finance to entertainment and education. While structured query languages (SQL) provide database experts the resources to extract, manipulate, and reason over this data, many potential users remain cut off from direct access due to the steep learning curve of mastering these languages. The importance of making data more accessible and actionable for a wider audience cannot be understated, given the growing centrality of data-driven decision-making in modern society. The vision of natural language interfaces to databases (NLIDB) is rooted in this very imperative—to allow non-experts to interact with databases using familiar, everyday language.  This reinforces the importance of developing modern text-to-SQL systems that can also reason over databases.

\begin{figure}[t]
    \centering
    \includegraphics[width=\linewidth]{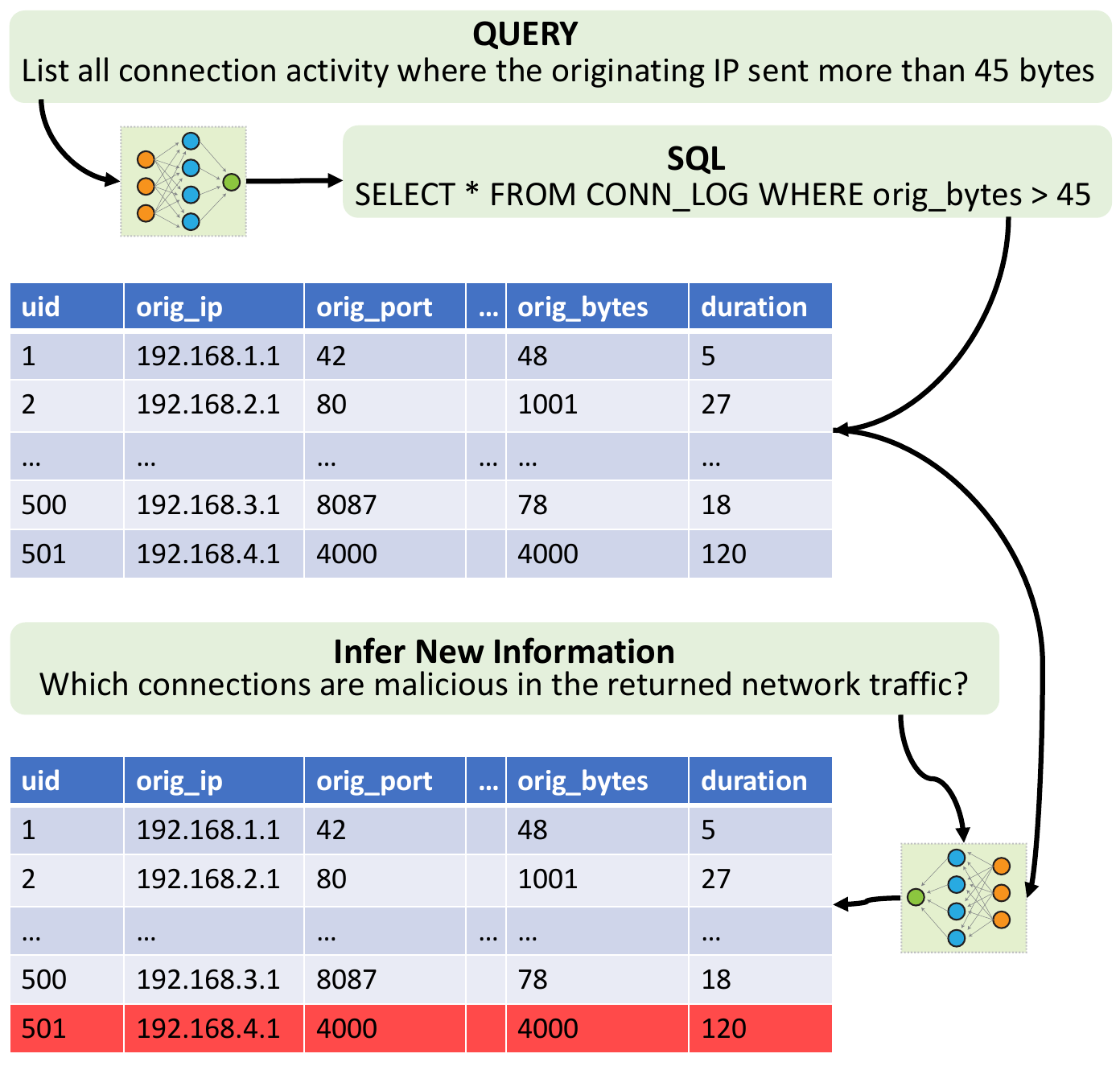}
    \caption{This figure provides an example of querying and reasoning over network traffic data.}\vspace{-2em}
    \label{fig:example}
\end{figure}

A system that seamlessly translates natural language queries into SQL (text-to-SQL) not only democratizes access to data but also has the potential to drastically reduce the time to insights for diverse stakeholders, including managers, analysts, educators, and the general public. There have been many advances in translating natural language to SQL~\cite{xu2017sqlnet,zhong2017seq2sql,bogin2019representing,wang2018robust,yu2018typesql,scholak2021picard,xie2022unifiedskg,wang2022proton,chen2021evaluating,sun2022leveraging}. Recent work has focused on either fine-tuning transformers or on the use of pre-built large language models (e.g., ChatGPT) with prompt tuning and in-context examples. For example, \citet{pourreza2023din} explored in-context learning using ChatGPT to generate SQL statements, and  
\citet{dong2023c3} explored zero-shot text-to-SQL generation using ChatGPT. \citet{wang2020rat} developed a unified framework using fine-tuning for text-to-SQL generation, leveraging relation-aware self-attention, to tackle schema encoding, schema linking, and feature representation. Combined with BERT data augmentation, this framework yielded a remarkable exact match accuracy of 65.6\% on the Spider dataset.

Much of the prior work on text-to-SQL generation has focused on simply generating SQL statements from the input text queries. Some recent work has expanded on standard studies by exploring conversational text-to-SQL tasks~\cite{yu2019cosql}. Intuitively, \citet{yu2019cosql} developed a system that can ask follow-up questions to answer ambiguous queries better, verify returned results, and notify users of unanswerable queries. However, there is limited work that can query a database and make inferences (understand) the returned data. Follow-up questions may involve making inferences and returning results that are not directly within the database. Hence, translating natural language to SQL is only half the challenge. The true power of such a system lies in its ability to retrieve and infer new information about the data returned. This ensures that the insights drawn from databases are accurate and meaningful. For instance, in an educational context, a student might not only ask for the number of historical events in a given time but might also want to know their significance or interconnections, requiring a depth of reasoning beyond retrieval.

At a high level, our work combines two lines of research not explored in previous papers: tabular data classification and question answering using transformers~\cite{badaro2023transformers} and text-to-SQL generation. There has been some recent work about predicting various aspects of tabular data. For example, \citet{yang2021exploring} predicts whether a claim is true or false given an input table. Likewise, \citet{deng2022turl} developed a system to inform missing or corrupted data within a table. However, much of this work assumes the table is provided. Hence, we develop a new text-to-SQL dataset to make predictions/inferences about the data and query the data using a single model. An example of our task is provided in Figure~\ref{fig:example}. As a case study, our dataset consists of Internet-of-Things (IoT) data from a smart building setting. Specifically, we assume a centralized database that captures both network traffic about the IoT devices and sensor readings (temperature, humidity, CO2 levels, etc.). The SQL statements query the IoT databases to return relevant data. The reasoning component of our dataset is specific to the network data. We classify the network traffic as malicious (e.g., DDoS attacks, botnet activity, etc.) or benign (non-malicious activity). Our decision to use IoT data is due to the following reasons. First, IoT data has a huge temporal component~\cite{acar2020peek}. There have been limited text-to-dataset resources that contain many temporal-related queries (e.g., Spider is based on SQLite databases and does not support datetime columns). Second, making inferences about network traffic data is non-trivial and has not been explored in the NLP community.

In summary, the contributions of this paper are as follows:
\begin{enumerate*}[label=\bfseries(\roman*)]
    \item We introduce a new IoT-SQL dataset containing 10,985 unique text-SQL pairs and 239,398 rows of network traffic activity from Zeek logs with annotations for malicious and non-malicious activity (e.g., DDoS attacks). This dataset provides a new test bed for text-to-SQL models and LLMs towards both querying data an actually understanding it. Specifically, current state-of-the-art LLMs GPT3.5 fail to perform well on this dataset for the reasoning component.\footnote{The dataset will be released upon acceptance.}
    \item We evaluate the performance of text-to-SQL models that can jointly query and reason about the data (i.e., predict whether specific network traffic is malicious). Our results suggest that modeling both tasks together substantially improves text-to-SQL performance with limited impact on network-traffic malicious activity detection.
    \item We perform error analysis and provide examples of how jointly training to query and understand the data improved SQL generation.
\end{enumerate*}


\section{Related Work}

\vspace{1mm} \noindent \textbf{Text-to-SQL Datasets.}
Recent momentum has grown in evaluating text-to-SQL systems, especially their generalizability, with less focus on the medical domain. Text-to-SQL translates text into machine-readable formats. Several datasets exist for this task: ATIS~\cite{data-atis-original,data-atis-geography-scholar} (airline queries), Geography~\cite{data-geography-original,data-atis-geography-scholar} (geographical data), Restaurants~\cite{data-restaurants, data-restaurants-logic, data-restaurants-original} (restaurant details), WikiSQL~\cite{zhong2017seq2sql}, Spider~\cite{yu2018spider}, and IMDB and Yelp \cite{data-sql-imdb-yelp} (movie and business data). The Spider dataset emerges as a cornerstone resource in the text-to-SQL benchmarks landscape. Designed to evaluate text-to-SQL systems rigorously, Spider boasts impressive extensiveness and diversity, featuring over 10,000 questions from over 200 databases. Its strength lies in its volume and the complexity of its queries. 

Recent efforts have also been made to develop new datasets beyond traditional text-to-SQL pairs.  \citet{yu2019cosql}, for example, collected a conversation-like corpus where a system can ask follow-up questions to answer ambiguous queries better, verify returned results, and notify users of unanswerable queries. Similarly, researchers have also focused on curating data (text-SQL pairs) that capture items missing in previous datasets (e.g., temporal-related queries). For example, \citet{vo2022tackling} introduced a new dataset called TempQ4NLIDB that contains 389 temporal-related question-SQL pairs to overcome limitations in existing datasets (e.g., Spider). Our research expands on this work, containing more than 1,000 temporally-related queries using MySQL datetime columns.

\vspace{1mm} \noindent \textbf{Text-to-SQL Methods.}
The field of text-to-SQL is concerned with automatically translating natural language queries into structured SQL queries. Recent advancements in neural network models have led to significant improvements in the accuracy and efficiency of Text-to-SQL systems~\cite{xu2017sqlnet,zhong2017seq2sql,bogin2019representing,wang2018robust,yu2018typesql,scholak2021picard,xie2022unifiedskg,wang2022proton,chen2021evaluating,sun2022leveraging}. 

Recent work has focused on fine-tuning transformers or using pre-built large language models (e.g., ChatGPT) with prompt tuning and in-context examples. For example, \citet{pourreza2023din} explored in-context learning using ChatGPT to generate SQL statements, and  
\citet{dong2023c3} explored zero-shot text-to-SQL generation using ChatGPT. \citet{wang2020rat} also proposed a relation-aware self-attention mechanism for text-to-SQL generation, achieving an accuracy of 65.6\% on the Spider dataset when combined with BERT~\cite{wang2020rat}. In another independent work, \citet{scholak2021picard} introduced the PICARD method, which uses incremental parsing for fine-tuning formal languages. This led to state-of-the-art results on both the Spider and CoSQL datasets. \citet{wang2022proton} introduced a novel approach to schema linking using the Poincaré distance metric. Their results established a new benchmark in performance, outperforming rule-based methods across multiple datasets and showcasing the effectiveness of their probing method. A more recent thorough analysis of the Codex language model's text-to-SQL abilities was undertaken by \citet{rajkumar2022evaluating}, whose findings highlighted the model's competitive performance across benchmarks, even without finetuning. Particularly on the Spider benchmark, Codex achieved an accuracy of up to 67\%. Their work also indicated that using a small set of in-domain examples could boost Codex's performance beyond some finetuned state-of-the-art models.

\vspace{1mm} \noindent \textbf{Tabular Data Understanding.}
There has been a wide array of papers about understanding tabular data beyond text-to-SQL~\cite{badaro2023transformers}. According to \citet{badaro2023transformers}, there are six common tabular data tasks: Fact-checking, question answering, semantic parsing (i.e., text-to-SQL), table retrieval, table metadata prediction, and table content population. Fact-checking related work has generally focused on predicting whether a statement/claim is factual, given the knowledge available in a Table~\cite{yang2021exploring}. Table retrieval research has focused on finding a table that contains the answer to a particular question~\cite{wang2022table,wang2021retrieving}. Table metadata prediction involves predicting information about the table, such as the column name or a relation between two columns~\cite{suhara2022annotating,du2021tabularnet}. Finally, table content population involves filling the cells within a table because of missing or incorrect data~\cite{iida2021tabbie,tang2021rpt}.

Intuitively, our task can be considered a combination of semantic parsing and table content population. The former (semantic parsing) is the text-to-SQL task, and the table population we are predicting is malicious or benign information for network traffic. We can think of the malicious information as a missing column in the database. But, more importantly, this is a highly specialized task that large language models cannot easily reason about. Hence, our dataset provides a unique research test bed for integrating highly specialized knowledge into LLMs for tabular QA.

\section{Data}

\begin{figure*}
    \centering
    \includegraphics[width=\linewidth]{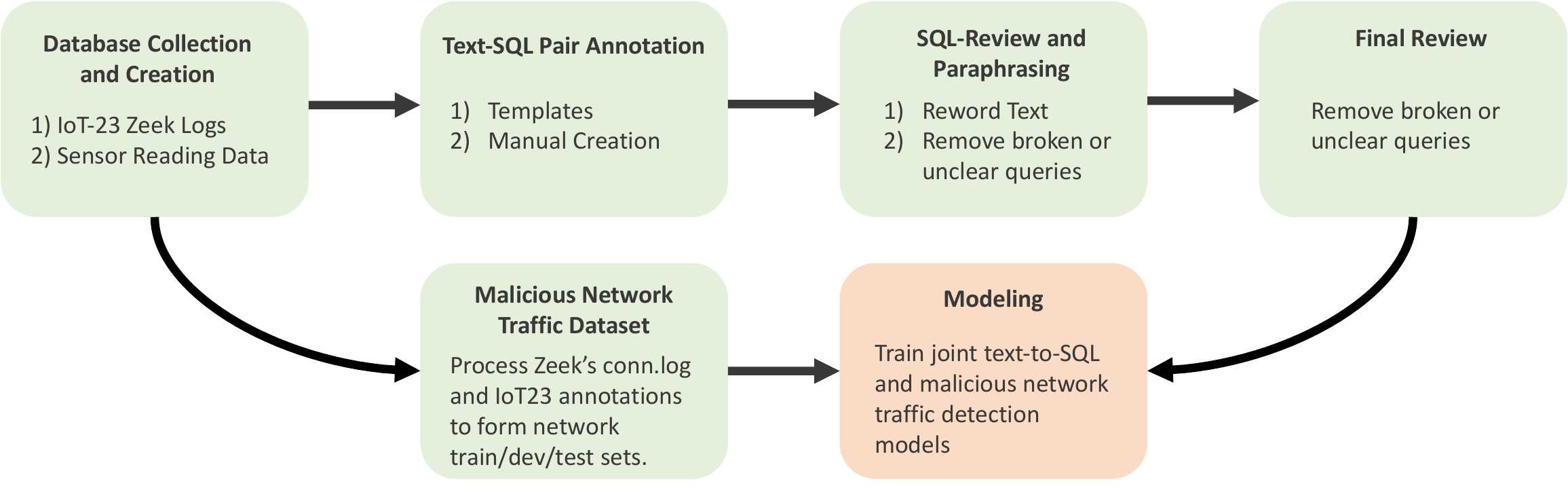}
    \caption{Text-to-SQL and malicious network traffic data collection pipeline overview.}\vspace{-1em}
    \label{fig:pipeline}
\end{figure*}

\begin{table}[t]
\renewcommand{\arraystretch}{1.2}

\centering
\resizebox{.8\linewidth}{!}{%
\begin{tabular}{lrrr}
\toprule
 & Train & Dev & Test  \\ \midrule
\# Examples & 6591 & 2197 & 2197  \\
Average Question Length & 2.3 & 2.3  & 2.5 \\
Min Question Length & 5 & 6 & 6  \\
Max Question Length & 63 & 53 & 46  \\
Average SQL Length & 16.3 & 16.5 & 16.4  \\
Min SQL Length & 5 & 5 & 5   \\
Max SQL Length & 146 & 140 & 140  \\ \midrule
\# Tables & \multicolumn{3}{c}{12} \\
\# Columns & \multicolumn{3}{c}{173} \\ \bottomrule
\end{tabular}%
}
\caption{Basic overview of the the text-to-SQL data.} \vspace{-1em}
\label{tab:sqlstats}
\end{table}

\begin{table}[t]
\centering
\renewcommand{\arraystretch}{1.2}

\resizebox{.8\linewidth}{!}{%
\begin{tabular}{@{}llll@{}}
\toprule
 & Train & Dev & Test \\ \midrule
\# Examples & 125,000 & 57,199 & 57,199  \\
\# Malicious Examples & 50,000 & 19,701 & 19,697  \\
\# Features & 19 & 19 & 19  \\ \bottomrule
\end{tabular}%
}
\caption{Basic overview of the network traffic data used to train and evaluate malicious traffic.}
\vspace{-1em}
\label{tab:netstats}
\end{table}

In this section, we describe the data creation process for text-SQL pairs, the source of the network traffic and sensor data, and how the network traffic data was organized for training our malicious network traffic activity detection model. As shown in Figure~\ref{fig:pipeline}, the data curation pipeline comprises five major steps. First, we curate the data for the database. Second, we ``annotate'' text-SQL pairs. Third, we partition network traffic data from the database to be used to train and evaluate a malicious traffic detector. Fourth, we review the text-SQL pairs, removing incorrect, irrelevant, or unclear queries. Moreover, we paraphrase each text-SQL pair to provide diversity in how things are specified. Finally, we perform an additional round of review after the paraphrase process.

\subsection{Database Collection and Creation}

We curate the data for our IoT database from two sources: IoT-23~\cite{garcia2021iot} and the Smart Building Sensor Data~\cite{hong2017high}.

\vspace{1mm} \noindent \textbf{IoT-23.} The IoT-23 dataset is created to facilitate the development and validation of intrusion detection systems (IDS) for IoT devices. It contains benign and malicious network traffic recordings. The network traffic recorders are stored in PCAP files and Zeek logs. For this study, we focus on the Zeek logs. Zeek~\cite{paxson1999bro}, formerly known as Bro, is an open-source network security monitoring tool. Its primary purpose is to analyze network traffic and generate high-level logs, metrics, and events that abstract the raw data into more meaningful and actionable insights. Zeek is widely used in network security, monitoring, and forensic analysis. There are conn.log, dns.log, files.log, http.log, npt.log, and weird.log. The conn.log records connection-level information detailing the sessions seen on the network. A list of the columns in the conn.log is found in Table~\ref{tab:conn}. Each row in the conn.log is annotated with malicious or benign and the type of malicious activity (e.g., DDoS, command and control, specific malware, and more). We discuss this more in the Network Traffic subsection. dns.log contains DNS request and response data. files.log stores details about files transferred over supported protocols, such as HTTP or FTP. http.log captures detailed HTTP request and response information. ntp.log contains information related to NTP transactions, such as timestamp updates, server-client interactions, version details, and other attributes specific to NTP communications. Finally, weird.log logs anomalies or unusual behaviors in network traffic. Each dataset is processed and stored as an independent table in the database.\footnote{More details on Zeek logs can be found at \url{docs.zeek.org/}}


\vspace{1mm} \noindent \textbf{Smart Building Sensors.} The Smart Building Sensor Data is a dataset derived from 255 sensors strategically deployed across 51 distinct rooms spanning four floors of a university building. The dataset contains humidity, CO2, temperature, luminosity, and motion sensor readings. Each reading is related to a specific room in the building. This dataset presents a unique opportunity for empirically exploring patterns associated with indoor spaces' physical attributes, particularly when combined with network traffic in a synthetic building-level database. Each sensor type (humidity, luminosity, etc.) is stored as a unique Table in our database, where each row represents a sensor read. Intuitively, the goal is to have a comprehensive database that may be used in a smart building setting, containing both the raw sensor information and meta data (network traffic) for smart devices.

\subsection{Text-to-SQL Pair Annotation}
The SQL queries were created using two major approaches: programmatically using a templated approach similar to \citet{wang2020text} and manually creating text-SQL pairs without templates. We describe each of these approaches in detail below:


\vspace{1mm} \noindent \textbf{Templates.}
Following the work by \citet{wang2020text}, we generate templates that fit two categories: retrieval queries and reasoning queries. Retrieval queries are primarily meant to extract specific records or data from the database. Reasoning queries are more complex and often involve several logical operations and conditions. They often require the model to comprehend intricate relations between different parts of the question or between multiple database tables. The distinction is helpful because different query types can be challenging in their ways. Retrieval queries test the model's ability to correctly identify and fetch data, while reasoning queries test its ability to process and integrate multiple pieces of information.

In total, we created 27 templates containing simple and complex queries. Templates are generated to create queries containing JOINs, HAVING statements, aggregation operations (e.g., average), and nested queries.\footnote{Code and a comprehensive list of all the queries will be made upon acceptance.}. An example template is
\begin{verbatim}
 SELECT $AGG_OP ($AGG_COLUMN)+
  FROM $TABLE WHERE ($COND_COLUMN
  $COND_OP $COND_VALUE)+
\end{verbatim}
In the above expression, \verb|$AGG_OP| represents aggregation methods (e.g., AVG(), MAX(), and MIN()), \verb|$AGG_COLUMN| represents the column to perform the aggregation on (e.g., ``duration'' from conn.log), \verb|$TABLE| represents the table the column is pulled from, \verb|$COND_COLUMN| (e.g., orig$\_$h representing the IP address), \verb|$COND| represents a conditional operator (e.g., $>$, $<$, $=$), and \verb|$COND_VALUE| represents the value to check (e.g., 192.168.1.1). An example query generated from the template is
\begin{verbatim}
 SELECT AVG(duration)
  FROM CONN_LOG WHERE (orig_h
  = "192.168.1.1")
\end{verbatim}
where items such as \verb|$AGG_OP| are replaced with \verb|AVG()|.

After creating the text-SQL pairs using templates, we paraphrased (reworded) each text piece to add diversity in the ways each question type is asked. Six researchers manually paraphrased each question. For instance, the automatically generated sentence, ``\textit{List the distinct proto for the DNS LOGs table with TTLs equal to 2523}''
would be transformed into ``\textit{Provide a list of unique DNS proto values with a TTLs value of 2523}'',
where the sentence is now more natural. All students had expertise in databases and were data analytics majors. The text-SQL pairs were assigned randomly to each researcher. In total, we create a total of 10,000 text-SQL pairs using templates.

\vspace{1mm} \noindent \textbf{Manual Creation.}
It is difficult to create templates that capture complex or unique queries. Hence, student researchers also manually created text-SQL pairs without using template-generated pairs. In total, 985 manually curated pairs were collected.

\vspace{1mm} \noindent \textbf{SQL-Review and Dataset Statistics.} After curating and paraphrasing the text-SQL pairs, we performed a multi-round review process. Each text-SQL pair was reviewed to measure whether the text was clear. This was done by having different annotators review another annotator's text-SQL pairs and paraphrases to ensure they could create the same SQL statement. Each researcher would create an SQL prompt, test the logic against a database, and after the query is successfully executed, SQL questions would be generated from the tables and variables in the Iot-23 dataset. Also, there were situations where manual text-SQL pairs were either incorrect or unrealistic; hence, these pairs were removed or paraphrased before incorporating them into the entire dataset. Overall, \textbf{the entire data collection process took 1.5 years.} The final annotated data statistics can be seen in Table~\ref{tab:sqlstats}. The dataset used to train the text-to-SQL models consisted of 10,985 rows. Each row contained a SQL query and a corresponding description, question, or prompt. The SQL queries varied in complexity but consisted primarily of arguments such as select distinct, max, avg, having, filtering, and join. On average, the prompts contained sixteen words, with the shortest prompt containing five words and the longest containing 146. 

\subsection{Network data}
The network traffic data comes from the IoT-23 dataset, which is used to train and evaluate our ability to detect malicious activity. We split the data into train, validation, and test sets based on attack type. Each session in the conn.log is labeled with one of ten attack-related labels: Attack, {Benign}, {C\&C}, {DDoS}, {FileDownload}, {HeartBeat}, {Mirai}, {Okiru}, {Torii}, and {PartOfAHorizontalPortScan}. 

\citet{kus2022false} show that training and testing on the same attack types results in high F1 scores but performance drops for unknown attack types. To address this, we ensure the same attack type in the training set is not in the validation and test sets. The training set contains network traffic related to PartOfAHorizontalPortScan and Okiru, while other attack types are used in validation and test datasets. All malicious activities are merged into a single ``malicious'' label, and IP addresses and timestamps are randomized to prevent data leakage.
The appendix provides details of the columns and features used and a summary of the data.



\section{Method}

In this section, we describe the approach we developed to address the text-to-SQL task and malicious traffic detection tasks jointly.


\vspace{1mm} \noindent \textbf{Schema for text-to-SQL.} \label{schema}
The table schema must be included with the model input to train a model to generate SQL queries specific to our database. The schema includes all tables and variables from our database (IoT and sensor data).  Formally, let $ t_i $ represent a table $ i $, and let $ c_{i,j} $ represent a column $ j $ in table $ i $. Each column has an attribute $ a_{i,j} $ representing the $ j $-th column's datatype in table $ i $. For instance, we have the table conn.log, which stores information about connections/sessions. Two columns within conn.log include \texttt{orig$\_$h} and \texttt{orig$\_$p}. The attribute assigned to the \texttt{orig$\_$h} column is \texttt{text} since it contains strings (IP addresses). The attribute assigned to \texttt{orig$\_$p} is \texttt{number} (representing the port number). Given all of the tables, columns, and attributes in a database, we generate the schema represented in the form of $s = [*, t_1, c_{1,1}, a_{1,1}, c_{1,2}, a_{1,2}, t_2, c_{2,1}, a_{2,1}, \ldots]$. In practice, this looks like $s$ = [*, conn.log, orig$\_$p, text, $\ldots$, weird.log, orig$\_$p, text, $\ldots$]. We concatenate the schema to each input text before being passed to the T5 models to generate the SQL statements.

\begin{table*}[t]
\centering
\renewcommand{\arraystretch}{1.2}

\resizebox{.8\textwidth}{!}{%
\begin{tabular}{@{}llrrrr@{}}
\toprule
 &  & \multicolumn{2}{c}{\textbf{Validation}} & \multicolumn{2}{c}{\textbf{Test}} \\ \midrule
 &  & \textbf{Execution Acc} & \textbf{Logical Acc} & \textbf{Execution Acc} & \textbf{Logical Acc} \\ \cmidrule(lr){3-4} \cmidrule(lr){5-6}  \midrule
  &  \multicolumn{5}{c}{\textbf{Methods that can only Generate SQL Statements}} \\ \midrule

\multirow{3}{*}{\textbf{Fine-tuned}}  & \textbf{BART} & .693 & .233 & .400 & .232 \\ 
& \textbf{T5-base} & .904 & .729 & .827 & .746 \\
 & \textbf{T5-large} & .966 & .868 & .928 &  .861 \\ \midrule \midrule
 &  \multicolumn{5}{c}{\textbf{Methods that can detect Malicious Traffic and Generate SQL Statements}} \\ \midrule
  \multirow{2}{*}{\textbf{Prompt-based}} 
   & \textbf{GPT3.5 Few-Shot} & .813 & .147 & .841 & .177 \\ \midrule
\multirow{1}{*}{\textbf{Fine-tuned + Malware MT Learning}} & \textbf{T5-base} & .927 & .837 & .956 & .851 \\
 \bottomrule
\end{tabular}%
}

\caption{Text-to-SQL generation results}
\vspace{-1em}
\label{tab:sqlres}
\end{table*}

\vspace{1mm} \noindent \textbf{Input for Malicious Traffic Detection.} \label{malsch}
Instead of passing the schema and text as input for malicious traffic detection, as we do for the text-to-SQL generation, we pass an instruction and formatted tabular data. Let $p$ represent the instruction and $t$ represent the formatted tabular data. We concatenate both to form the input $x = [p, t]$. This work uses the instruction ``Is the following network information Malicious?''. Also, the tabular data (row) is formatted as $t_1 t_2 \cdots t_n$, where each tabular data column/value is represented as a string. Moreover, everything is concatenated using space as the delimiter. In practice, this looks like ``192.168.1.1 80 192.161.2.2 8080 $\cdots$.'' Note that there are no spaces in the values available in the conn.log file, which contains the network data used for malicious traffic detection. If this work is expanded to other Zeek logs, other delimiters would need to be explored.

\vspace{1mm} \noindent \textbf{Training.}
To train the model, we fine-tune the Flan-T5-base~\cite{chung2022scaling} model. The model is trained using the Adam optimizer~\cite{kingma2014adam} with a minibatch size 4 and a learning rate .0001. We trained the model for a total of 15 epochs. The model was trained by simply combining the data formatted as described in Sections~\ref{schema}.

\section{Results}

In this section, we describe the evaluation metrics, our baseline models, and the results for text-to-SQL prediction and malicious network traffic detection. We also provide an informative error analysis.

\vspace{1mm} \noindent \textbf{Evaluation Metrics.}
We use two primary metrics for evaluating the text-to-SQL results: \textit{Logical Accuracy} and \textit{Execution Accuracy}. Logical Accuracy assesses the correctness of the logical structure and semantics of the generated SQL with the target SQL (i.e., measuring whether two SQL queries are exactly the same). However, a potential pitfall of relying solely on Logical Accuracy is that two queries may be correct but written differently. On the other hand, Execution Accuracy evaluates the results obtained when the generated SQL is run on a database. This metric is vital because the ultimate goal is to extract accurate information from the database, regardless of the SQL's structure. However, a high Execution Accuracy doesn't guarantee that the SQL query is optimal or semantically correct. It's possible for an inefficient or technically incorrect query to yield the desired results that are returned by the ground-truth query. Hence, we consider both Logical and Execution Accuracy in our study. We use the standard classification metrics macro-precision, macro-recall, and macro-F1 to evaluate our models' malicious network traffic detection performance.

\begin{table*}[t]
\centering
\renewcommand{\arraystretch}{1.2}
\resizebox{.8\textwidth}{!}{%
\begin{tabular}{lrrrrrrr}
\toprule
 &  &  & \textbf{Validation} &  &  & \textbf{Test} &  \\ \midrule
 &  & \textbf{precision} & \textbf{recall} & \textbf{F1} & \textbf{precision} & \textbf{recall} & \textbf{F1} \\ \cmidrule(lr){3-5} \cmidrule(lr){6-8} \midrule
 &  \multicolumn{7}{c}{\textbf{Methods that can only detect Malicious Traffic}} \\ \midrule
\multirow{4}{*}{\textbf{Baselines}} & \textbf{Stratified} & .500 & .500 & .498 & .502 & .502 & .501 \\
 & \textbf{Uniform} & .503 & .503 & .491 & .497 & .497 & .485 \\
 & \textbf{Random Forest} & .879 & .697 & .714 & .878 & .694 & .710 \\
 & \textbf{SVM} & .874 & .693 & .709 & .872 & .689 & .704 \\ \midrule

\multirow{2}{*}{\textbf{Fine-tuned}} & \textbf{T5-base} & .883 & .708 & .728 & .882 & .704 & .723 \\
 & \textbf{T5-Large} & .900 & .777 & .804 & .904 & \textbf{.775} & \textbf{.802} \\ \midrule
 \midrule
&  \multicolumn{7}{c}{\textbf{Methods that can detect Malicious Traffic and Generate SQL Statements}} \\ \midrule
   \multirow{2}{*}{\textbf{Prompt-based}} &  \textbf{GPT3.5 Zero-Shot} & .167 & .388 & .215 & .183 & .392 & .220 \\
  & \textbf{GPT3.5 Few-Shot} & .741 & .761 & .711 & .671 & .640 & .543 \\ \midrule
\multirow{1}{*}{\textbf{Fine-tuned + Malware MT Learning}} & \textbf{T5-base} & .810 & .684 & .697 & .808 & .680 & .693 \\  \bottomrule
\end{tabular}%
}
\caption{Malicious traffic detection results.}
\vspace{-1em}
\label{tab:netres}
\end{table*}

\vspace{1mm} \noindent \textbf{Baseline models.}
We explore two major baselines to evaluate the performance of detecting malicious web traffic: Support Vector Machines (SVM) and Random Forest. The input of the models includes all of the features listed in Table~\ref{tab:conn} except for ts, uid, origh$\_$h, resp$\_$h, and tunnel\_parents (i.e., all unique identifiers and IP addresses are removed). The models used to create the baseline include stratified, uniform, random forest, and support vector machines. We also explore two random baselines: stratified and uniform. The stratified baseline randomly predicts each class based on the class proportion in the training dataset, and the uniform baseline randomly predicts each class with equal probability. Finally, we evaluate transformer models Flan-T5-base and Flan-T5-Large where the input is formatted as described in Section~\ref{malsch}. Finally, we evaluate using GPT3.5 with few-shot prompts (64 examples). For the GPT3.5 model, the data is supplied in a json-like format (label, value) pairs so it knows what each value represents. 

For text-to-SQL, we explore two types of fine-tuned baselines. For the fine-tuned models for text-to-SQL, we evaluate three models: Flan-T5-base, BART~\cite{lewis2020bart}, and Flan-T5-large. Each model is trained using the same schema defined in Section~\ref{schema}. These models are not trained on the network traffic data. We also evaluate GPT3.5 using in-context examples. We provide 64 in-context examples from the training dataset to make predictions. In general, our GPT3.5 prompt follows the work of \citet{gao2023text}, which achieved state-of-the-art performance on the Spider dataset~\cite{yu2018spider}.

\vspace{1mm} \noindent \textbf{Text-to-SQL.}
In Table~\ref{tab:sqlres}, we report the results on the text-to-SQL task. We compare the baselines to models fine-tuned only on the text-to-SQL corpus and to a model trained on the text-to-SQL and network traffic data. Overall, we find that the larger model T5-Large outperforms the T5-base model when fine-tuned only on text-to-SQL data. The T5-Large model achieves a logical accuracy of .861 on the test set and an execution accuracy of .928. However, when jointly trained on both datasets, we find that the T5-base model can nearly match (and beat) the performance of the larger model. Specifically, the T5-base model achieves a logical accuracy of .851 and an execution accuracy of .956 on the test data with multi-task training, thus matching and outperforming the T5-Large model trained only on the text-to-SQL data.  

\vspace{1mm} \noindent \textbf{Malicious Network Traffic Detection.}
In Table~\ref{tab:netres}, we report the results of detecting malicious network traffic. We find that the Random Forest outperforms other methods for the baseline models. The random forest model had an F1 score of .710 and a recall score of .694. The SVM had similar results, with an F1 score of .704 and a recall score of .689. Moreover, the GPT3.5 method performs poorly on the task, with only an F1 of .543 on the test step, a light improvement over random. We hypothesize that the validation performance is slightly better because the LLM was able to understand those attacks better than the test set attacks. However, the transformer-based models (T5-base and T5-large) substantially outperformed all baseline models. The Flan-T5-Large model was the top-performing fine-tuned model model, with an F1 score of .802 and a recall score of .775. Overall, compared to the text-to-SQL results, we find that training on both malicious traffic detection and text-to-SQL reduces the performance of the network traffic task. When analyzing the results, we find that the model struggles to identify malicious items, mostly labeling examples as Benign. 

\vspace{1mm} \noindent \textbf{Error analysis.}
Why does the T5-base model match and outperform the T5-large model when trained on both datasets? Our analysis shows that much of the improvement is on the conn.log-related queries. The conn.log was the table used as input when training the malicious network traffic detection-related aspect of our model. Specifically, for logical accuracy, 142 examples in the test dataset contained items related to the conn.log table. The T5-base model missed 42 of them. The jointly trained T5-base model only missed 27. Some of the errors were major, where the T5-base model did not generate a SQL statement at all, where the T5-large model returned the correct statement (e.g., ``SELECT service FROM IoT23$\_$CONN$\_$LOG GROUP BY service HAVING AVG(resp$\_$bytes) >= 829'').

We also hypothesize that while we did not train to make inferences about other tables in the database, by better understanding the conn.log table, the model can better understand how it relates more to other tables via JOIN queries. This better understanding of table relationships potentially results in improvements for other tables as well.

\section{Conclusion}

Databases hold large amounts of structured knowledge across various sectors, and efficient access to this data is essential. Our study was driven by the goal of NLIDB, which is to simplify data access beyond the complexities of SQL. While there have been advancements in text-to-SQL systems, our research emphasizes the importance of retrieving and understanding the data. With the introduction of the IoT-SQL dataset, we've provided a unique resource with the ability to predict aspects not in the database (i.e., malicious network activity) and generate SQL statements based on an input text query. Moreover, the dataset contains many temporal queries that are missing or limited in prior text-to-SQL datasets. Our findings show that models trained to query and reason about data improve SQL generation performance. 

Overall, there are two major avenues for future work. First, we plan to explore more complex models on the dataset, particularly on more complex training, validation, and test sets. For example, recent work suggests that exploring different data split methods (e.g., based on SQL length, tables, or column names) can improve the measure of generalizability~\cite{gan2022measuring,tarbell2023towards}. Second, we will explore more sophisticated methods of detecting malicious network activity. Malicious activity may be related to multiple sessions within the Zeek Conn.log. Developing a system that can reason over multiple rows in the database can potentially generate substantial improvements.

\section{Acknowledgements}
We acknowledge Dr. Glenn Dietrich, a co-author who passed away before this paper was published. This work would not have been possible without his support.

\section{Limitations}

Our study acknowledges several limitations that warrant discussion. Firstly, while our novel IoT-SQL dataset provides a rich collection of text-SQL pairs and network traffic data, the specific focus on IoT environments and network traffic may limit the generalizability of our findings to other domains or types of data. This specialization means that models trained on our dataset might not perform as well when applied to databases with different structures or content, such as healthcare or financial databases. However, it is still a novel domain for tabular QA, which state-of-the-art LLMs (e.g., GPT3.5) struggle to understand, thus providing a new testbed for understanding how to add new functionality to the models. We also understand that GPT4 may perform better than GPT3.5, but because of the size of the network data, the experiments are expensive. GPT3.5 experiments cost nearly \$600, not including small preliminary experiments. There are also things that could have improved the results, e.g., finding the most similar in-context examples. But, again, the cost was prohibitive because of our limited research budget.

Also, our approach relies heavily on the quality and diversity of the SQL queries and the paraphrased text. Despite our efforts to generate diverse and complex queries, certain query structures or linguistic variations may still be underrepresented. This underrepresentation could impact the model's ability to generalize across unseen queries or to handle nuanced variations in natural language.

Another significant limitation lies in the multi-task learning approach for joint training on text-to-SQL generation and malicious network traffic detection. While this approach improved the text-to-SQL performance, it did not enhance and, in some cases, slightly reduced the accuracy of malicious traffic detection. This suggests a potential trade-off when balancing multiple tasks, and further research is needed to optimize such multi-task learning frameworks to ensure that improvements in one task do not come at the expense of another.

In summary, while our contributions are significant, addressing these limitations through future research will be crucial for advancing the state of text-to-SQL systems and their application to diverse and complex datasets to really understand all types of data beyond just generating a SQL statement.

\bibliography{lrec-coling2024-example}

\appendix

\section{Data}

\begin{table}[h]
\centering
\resizebox{0.8\linewidth}{!}{%
\begin{tabular}{lp{5cm}}
\toprule
IoT Data & Description \\ \midrule
ts &  Timestamp of the first packet\\
uid & Uniqie ID of the connection \\
id.orig\_h & Originating endpoint's IP address (Orig) \\
id.orig\_p &  Originating endpoint's TCP/UDP port (or ICMP code)\\
id.resp\_h & Responding endpoint's IP address (Resp) \\
id.resp\_p &  Responding endpoint's TCP/UDP port (or ICMP code)\\
proto & Transport layer protocol of connection \\
service &  Detecting application protocol, if any\\
duration &  Connection length\\
orig\_bytes &  Orig payload bytes, from sequence numbers if TCP\\
resp\_bytes &  Resp payload bytes; from sequence numbers if TCP\\
conn\_state &  Connection state \\
local\_orig &  is Orig in Site::local$\_$nets?\\
local\_resp &  is Resp in Site::local$\_$nets?\\
missed\_bytes & Number of bytes missing due to connection gaps \\
history & Connection state history \\
orig\_pkts& Number of Orig packets \\
orig\_ip\_bytes & Number of Orig IP bytes (via IP total$\_$length header field) \\
resp\_pkts &  Number of Resp packets \\
resp\_ip\_bytes & Number of Resp IP bytes (via IP total$\_$length header field) \\
tunnel\_parents & if tunneled, connection UID of encapsulating parent(s) \\ \bottomrule
\end{tabular}%
}
\caption{This table contains a description of the Zeek Connection log columns, which are used as features when predicting malicious activity.}
\label{tab:conn}
\end{table}

\subsection{Network data}
As previously discussed, the network traffic data originates from the IoT-23 dataset. A sample of the data is used to train and evaluate the performance of our ability to detect malicious activity. We split the network traffic data into train, validation, and test sets based on the attack type. Each session (row) in the conn.log is labeled with one of ten attack-related labels. An \textbf{Attack} label involves the infected device exploiting a vulnerable service on another system, like brute-forcing logins. \textbf{Benign} connections display no malicious intent. \textbf{C\&C} signifies a device's connection to a Command and Control server, observed through periodic communications or suspicious downloads. \textbf{DDoS} denotes the device's role in overwhelming a target by sending excessive traffic. \textbf{FileDownload} infers a device downloading potential threats based on connection sizes and endpoints. \textbf{HeartBeat} marks periodic, minimal exchanges with a C\&C server, ensuring active monitoring. \textbf{Mirai}, \textbf{Okiru}, and \textbf{Torii} are labels pointing to specific botnet attack patterns, with the latter two being less common than Mirai. Finally, \textbf{PartOfAHorizontalPortScan} identifies efforts to scan various systems on the same port for vulnerabilities.

Recent work exploring malicious network traffic detection has analyzed why much of the reported results are greater than 99\% F1~\cite{kus2022false}. A major cause for these results is the training and testing on the same attack types. When the attack type is unknown (i.e., zero-days), performance is not as high. Hence, we split the data into training and test/validation datasets so that the same attack type in the training dataset is not in the validation and test sets. The training dataset contains network traffic related to PartOfAHorizontalPortScan and Okiru. The other sessions from the conn.log with different attach types are used in other validation and test datasets. Next, we merge all malicious activity into a single ``malicious'' label. Moreover, to avoid potential data leakage, all IP addresses and time stamps were randomized when training and evaluating the malicious traffic detection models. A summary of the data used for training and evaluating the malicious network activity models is shown in Table~\ref{tab:netstats} and the columns/features are shown in Table~\ref{tab:conn}.

\end{document}